\documentclass[conference]{IEEEtran}
\IEEEoverridecommandlockouts

\usepackage{amsmath}
\usepackage{latexsym}
\usepackage{amssymb}
\usepackage{epsfig}
\usepackage{moreverb}
\usepackage{rotating}
\usepackage{enumerate}
\usepackage{graphics, graphicx,wrapfig}
\usepackage{fancybox}
\usepackage{float}
\usepackage{url}
\usepackage{tabularx}
\usepackage{array}
\usepackage{booktabs}
\usepackage{multirow}
\usepackage{listings}
\usepackage{hyperref}
\usepackage{array}
\usepackage{balance}
\usepackage{subcaption}
\usepackage{algpseudocode}
\usepackage{algorithm}

\algrenewcommand\algorithmiccomment[1]{\hfill$\triangleright$~#1}
\algrenewcommand\textproc{\textit} 

\def\BibTeX{{\rm B\kern-.05em{\sc i\kern-.025em b}\kern-.08em
    T\kern-.1667em\lower.7ex\hbox{E}\kern-.125emX}}
\begin{document}

\title{\textit{ColNet}: Collaborative Optimization in Decentralized Federated Multi-task Learning Systems}

\author{
    \IEEEauthorblockN{Chao Feng\IEEEauthorrefmark{1}, Nicolas Fazli Kohler\IEEEauthorrefmark{1}, Zhi Wang\IEEEauthorrefmark{1}, Weijie Niu\IEEEauthorrefmark{1}, Alberto Huertas Celdrán\IEEEauthorrefmark{1,2}, \\ G\'er\^ome Bovet\IEEEauthorrefmark{3}, Burkhard Stiller\IEEEauthorrefmark{1}}
    \IEEEauthorblockA{\IEEEauthorrefmark{1}Communication Systems Group, Department of Informatics, University of Zurich, 8050 Zürich, Switzerland \\{[cfeng, niu, huertas, stiller]}@ifi.uzh.ch, [nicolasfazli.kohler, zhi.wang]@uzh.ch}
    \IEEEauthorblockA{\IEEEauthorrefmark{2}Department of Information and Communications Engineering, University of Murcia, 30100--Murcia, Spain}

    \IEEEauthorblockA{\IEEEauthorrefmark{3}Cyber-Defence Campus, armasuisse Science \& Technology, 3602 Thun, Switzerland gerome.bovet@armasuisse.ch}
}

\DeclareRobustCommand*{\IEEEauthorrefmark}[1]{%
  \raisebox{0pt}[0pt][0pt]{\textsuperscript{\footnotesize #1}}%
}

\maketitle

\begin{abstract}
The integration of Federated Learning (FL) and Multi-Task Learning (MTL) has been explored to address client heterogeneity, with Federated Multi-Task Learning (FMTL) treating each client as a distinct task. However, most existing research focuses on data heterogeneity (e.g., addressing non-IID data) rather than task heterogeneity, where clients solve fundamentally different tasks. Additionally, much of the work relies on centralized settings with a server managing the federation, leaving the more challenging domain of decentralized FMTL largely unexplored. Thus, this work bridges this gap by proposing \textit{ColNet}, a framework designed for heterogeneous tasks in decentralized federated environments.

\textit{ColNet} partitions models into a backbone and task-specific heads, and uses adaptive clustering based on model and data sensitivity to form task-coherent client groups. Backbones are averaged within groups, and group leaders perform hyper-conflict-averse cross-group aggregation. Across datasets and federations, \textit{ColNet} outperforms competing schemes under label and task heterogeneity and shows robustness to poisoning attacks.
\end{abstract}

\section{Introduction}
\label{introduction}
The integration of Federated Learning (FL) and Multi-Task Learning (MTL) has gained significant attention for its ability to tackle diverse but related tasks in a collaborative and privacy-preserving manner. FL trains models across clients by exchanging model updates rather than raw data, preserving data privacy \cite{kairouz2021advances}. Decentralized FL (DFL) eliminates the need for a central server, enhancing system robustness by avoiding single points of failure \cite{beltran2023decentralized}. MTL, traditionally performed on a single machine, leverages task interdependence to improve generalization, reduce overfitting, and address data sparsity \cite{caruana1997multitask,zhang2021survey}. Extending MTL to federated settings results in Federated Multi-Task Learning (FMTL) \cite{smith2017federated}, combining FL’s privacy preservation with MTL’s collaborative task-sharing benefits.

In both centralized and decentralized FMTL, existing research primarily addresses the challenge of non-independent and identically distributed (non-IID) data across clients, a core issue in FL \cite{zhao2018federated}. Non-IID data arises in various forms, including attribute skew (differences in feature distributions), label skew (differences in label distributions), and temporal skew (time-based changes in data distributions), each complicating model training \cite{zhu2021federatedlearningnoniiddata}. However, while much progress has been made in handling non-IID data, most studies focus on label heterogeneity rather than deeper task heterogeneity. Many approaches assume that all clients access the same set of class labels, even in non-IID settings. In reality, clients may encounter only subsets of labels relevant to their tasks \cite{aoki2022heterogeneous}, creating disparities in outputs and limiting collaborative learning. Addressing task heterogeneity, where clients solve fundamentally different problems, remains a significant challenge, especially in decentralized settings.

Aggregation algorithms, like FedAvg, struggle in multi-task learning scenarios due to their reliance on model homogeneity \cite{huang2022learn}. Task heterogeneity introduces complexities, as clients may require distinct model architectures for tasks (\textit{e.g., }classification versus point detection), leading to conflicting and dominant gradients during training \cite{lu2024fedhca2,yu2024unleashing}. While recent methods such as FedBone \cite{fedbone} and FedHCA\(^2\) \cite{lu2024fedhca2} mitigate these challenges, they focus on centralized settings (CFMTL), leaving decentralized FMTL (DFMTL) largely unexplored. Decentralization amplifies these challenges due to the absence of a coordinating server, demanding robust methods for managing tasks and model diversity.

This work proposes \textit{ColNet}, a decentralized framework for training multi-task models in heterogeneous environments. \textit{ColNet} partitions models into a backbone and task-specific heads. \textit{ColNet} uses adaptive clustering that leverages model sensitivity and data sensitivity to automatically partition nodes into task-coherent groups. Within each group, backbone layers are aggregated using FedAvg. Group leaders then exchange the group-averaged backbones across groups through a hyper-conflict-averse (HCA) aggregation scheme, which reduces gradient conflict and strengthens cross-task collaboration.

\begin{table*}[h]
\centering
\caption{Overview of Research in Federated Multi-Task Learning}
\label{tab:summary_fmtl}
\resizebox{\textwidth}{!}{%
\begin{tabular}{lllll}
\toprule
\textbf{Federation Schema} & \textbf{Techniques} & \textbf{Research}  & \textbf{Approach used} & \textbf{Heterogeneity Addressed} \\ \hline
 & Convex & \textbf{MOCHA} \cite{smith2017federated} & Convex optimization, dynamic client inclusion  & Label heterogeneity \\
 & Optimization & \textbf{OFMTL} \cite{OFMTL}    & Convex optimization, dynamic client inclusion  & Label heterogeneity \\ \cmidrule{2-5}
 & Regularization & \textbf{Ditto} \cite{li2021ditto}& Bi-level optimization, local regularization& Label heterogeneity\\ 
 & & \textbf{PMTL} \cite{hu2021private}     & Differential privacy, noise-injected global aggregation & Label heterogeneity  \\  \cmidrule{2-5}
& Privatization & \textbf{FedPer} \cite{arivazhagan2019federated}   & Shared base layers, private personalization layers     & Label heterogeneity  \\ 
Centralized &  & \textbf{FedRep} \cite{collins2021exploiting} & Shared base layers, head freezing  & Label heterogeneity  \\  \cmidrule{2-5}
 & Grouping& \textbf{CFL} \cite{sattler2020clustered}& Client grouping, cosine similarity& Label heterogeneity   \\
 & & \textbf{HeurFedAMP} \cite{huang2021personalized}     & Dynamic grouping, cosine similarity       & Label heterogeneity   \\  \cmidrule{2-5}
 & Knowledge & \textbf{EFDLS} \cite{xing2022efficient}    & Knowledge distillation, local teacher-student models  & Label heterogeneity \\ 
 & Distillation& \textbf{FedICT} \cite{wu2023fedict}   & Two-way knowledge distillation, reduced model updates & Label heterogeneity   \\  \cmidrule{2-5}
& Hybrid& \textbf{FedBone} \cite{fedbone}   & Split learning, gradient projection, task adaptation  & Task heterogeneity   \\
 & & \textbf{FedHCA\(^2\)} \cite{lu2024fedhca2}  & Hyper conflict-averse aggregation, encoder-decoder architecture & Task heterogeneity\\ \hline
 & Regularization& \textbf{DCLPMN} \cite{vanhaesebrouck2016decentralized}   & Graph-based regularization, neighbor collaboration  & Label heterogeneity  \\ \cmidrule{2-5}
 Decentralized & Hybrid& \textbf{SpreadGNN} \cite{he2022spreadgnn}   & Decentralized aggregation, molecular data focus    & Label heterogeneity   \\ \cmidrule{2-5}
  & Hybrid& \textbf{\textit{ColNet}} (This work)   & Adaptive task clustering, shared backbone, hyper conflict-averse aggregation & Task heterogeneity  \\ \bottomrule
\end{tabular}
}
\end{table*}

Experiments on the CIFAR-10 and CelebA datasets demonstrate the effectiveness of \textit{ColNet}. Firstly, the framework's performance is analyzed with varying aggregation round frequencies, showing that more frequent aggregations improve model performance. Secondly, different levels of layer privatization are examined, revealing that sharing most layers benefits class label heterogeneity setups, while increasing privatization slightly enhances performance in task heterogeneity scenarios. Compared with other schemes, \textit{ColNet} consistently outperforms baselines in both class label and task heterogeneity scenarios. Additional robustness evaluation indicates that \textit{ColNet} is resilient to diverse poisoning attacks, including model poisoning and data poisoning, highlighting its capability to address diverse challenges in DFL.

\section{Related Work}
\label{related}
This section provides an overview of FMTL research, covering both centralized and decentralized approaches and comparing them based on the techniques employed and types of heterogeneity addressed, as presented in \tablename~\ref{tab:summary_fmtl}.

\textbf{A. Centralized Approaches}

FMTL was first introduced in 2017 with the MOCHA framework \cite{smith2017federated}, and since then, a variety of approaches have emerged to tackle different aspects of multi-task learning in centralized federated settings. Early work primarily focused on solving convex optimization problems, such as MOCHA and OFMTL \cite{OFMTL}, which were robust but unsuitable for modern non-convex deep learning methods. Subsequent research addressed this limitation, extending CFMTL to handle non-convex problems and introducing techniques like regularization, privatization, grouping, and knowledge distillation.

Ditto \cite{li2021ditto} introduced bi-level optimization techniques that allow each client to maintain a personalized model while regularizing it toward a global model. This approach ensures uniform performance across devices and enhances resistance to data and model poisoning attacks. Similarly, PMTL \cite{hu2021private} leverages relaxed differential privacy to learn personalized models while safeguarding client data. By adding noise during global aggregation, PMTL enables effective collaboration without compromising privacy. FedPer \cite{arivazhagan2019federated} tackled statistical heterogeneity by dividing models into shared base layers and private personalization layers. Extensions like FedRep \cite{collins2021exploiting} introduced mechanisms such as freezing the shared layers during early training phases to improve adaptability and achieve higher test accuracy.

Frameworks like CFL \cite{sattler2020clustered} introduced grouping strategies to cluster clients with similar data distributions, reducing the impact of conflicting gradients during aggregation. HeurFedAMP \cite{huang2021personalized} further refined this concept by using cosine similarity to dynamically determine grouping, allowing more granular adaptation to client data heterogeneity. EFDLS \cite{xing2022efficient} utilized knowledge distillation models to transfer knowledge efficiently within individual clients. This method enables collaborative learning by sharing only essential updates, reducing communication overhead. FedICT \cite{wu2023fedict} extended this idea by employing a two-way distillation process that customizes local models for specific tasks while maintaining a flexible and adaptive global model, making it suitable for diverse client environments.

Most existing studies primarily focus on class label heterogeneity, restricting their adaptability in dynamic environments. Frameworks like FedBone \cite{fedbone} introduced split learning to separate general models on a central server from task-specific models on clients. Similarly, FedHCA\(^2\) \cite{lu2024fedhca2} proposed innovative aggregation techniques, such as HCA Aggregation, to address gradient conflicts, but it was designed for centralized environments. These advancements have predominantly relied on centralized coordination, simplifying aggregation but introducing challenges like single points of failure, limited scalability, and privacy risks. 

\textbf{B. Decentralized Approaches}

DFMTL eliminates the need for a central server, allowing clients to communicate directly in a peer-to-peer manner. Despite the advantages, DFMTL introduces added challenges, such as managing gradient conflicts, asynchronous updates, and efficient client communication. Early research, such as DCLPMN \cite{vanhaesebrouck2016decentralized}, addressed these challenges by leveraging graph-based regularization to model task relationships, enabling collaboration between neighboring clients. SpreadGNN \cite{he2022spreadgnn} applied decentralized aggregation techniques tailored for tasks like molecular data analysis to improve scalability and address domain-specific requirements.

As presented in \tablename~\ref{tab:summary_fmtl}, while research on FMTL has made significant progress, most works address label heterogeneity and non-IID data rather than task heterogeneity, where clients solve different tasks. Additionally, decentralized settings remain underexplored compared to centralized frameworks, despite their potential for improved privacy and scalability. Existing frameworks rarely address task heterogeneity in decentralized environments, leaving a clear gap for Heterogeneous DFMTL. This work addresses this gap by proposing a novel framework, \textit{ColNet}, designed to handle both label and task heterogeneity in a decentralized setting.

\section{The \textit{ColNet} Solution}
\label{sec:solution}
This work introduces \textit{ColNet}, a collaborative optimization approach designed for multi-task learning in DFL. This section defines the DFMTL problem and outlines its optimization objectives. It then presents the proposed \textit{ColNet} framework, describing each step in turn.

\subsection{Problem Statement}
In a DFMTL system with $N$ nodes, the learning objective is to collaboratively learning task-specific models while preserving data privacy and operating under a fully decentralized communication pattern. 

Let \(\mathcal{V} = \{1, 2, \dots, N\}\) denote the set of nodes. Each node \(i \in \mathcal{V}\) is associated with a task \(\tau_i\) and holds a local dataset \(\mathcal{D}_i\). The number of distinct tasks is \(M \leq N\), accommodating the possibility that multiple nodes may share the same task or tasks may be heterogeneous across nodes.

The nodes form a connected, undirected graph \(G = (\mathcal{V}, \mathcal{E})\), where \(\mathcal{E}\) specifies the peer-to-peer (P2P) links. Communication is entirely decentralized, with no central server. Each node exchanges model updates or partial information solely with its neighbors \(\mathcal{N}(i)\).

Each node \(i\) maintains a parameter vector \(\mathbf{w}_i \in \mathbb{R}^d\) that addresses its local task \(\tau_i\). Heterogeneous local data render a single global model suboptimal, making personalized node-specific models more appropriate. Any common structure or correlation among tasks can be leveraged to improve overall performance.

Thus, the learning goal is framed as a joint optimization problem capturing both local performance and inter-task relationships:
\begin{equation}
    \min_{\{\mathbf{w}_i\}_{i=1}^N} \;
    \sum_{i=1}^N f_i\bigl(\mathbf{w}_i; \mathcal{D}_i\bigr)
    \;+\;
    R \bigl(\{\mathbf{w}_i\}_{i=1}^N \bigr),
    \label{eq:dfmt_objective}
\end{equation}
    where \(f_i(\mathbf{w}_i; \mathcal{D}_i)\) is the local loss function at node \(i\), and \(R(\{\mathbf{w}_i\})\) is a regularization or coupling term designed to exploit shared structure among tasks.

\subsection{\textit{ColNet} Architecture}
\begin{figure}[t]
    \centering
    \includegraphics[width=1\linewidth]{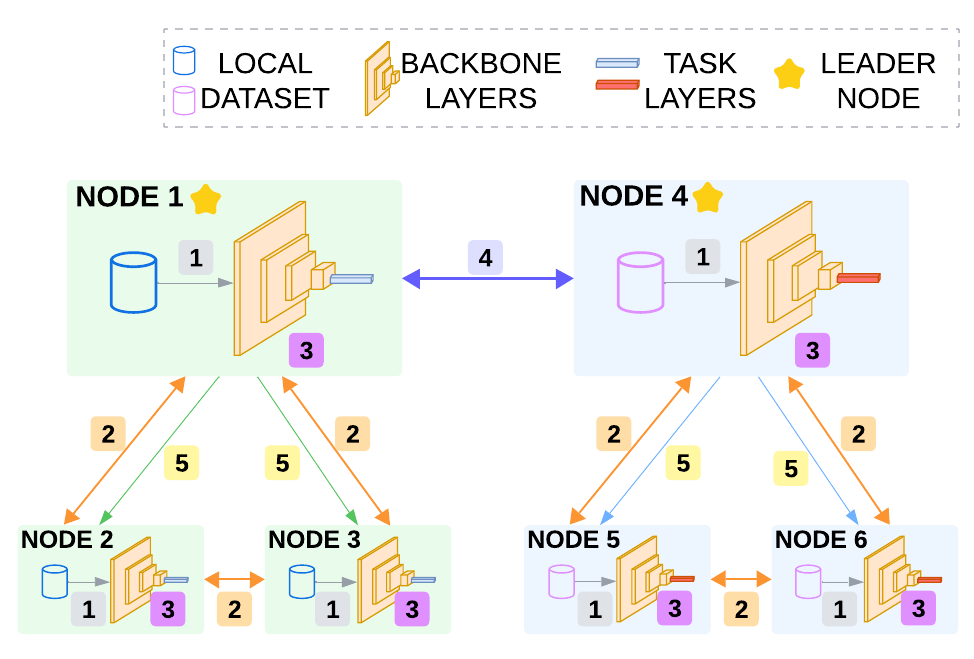}
    \caption{Overview of the \textit{ColNet} Learning Process}
    \label{fig:architecuture}
\end{figure}

To tackle the challenges of multi-task learning in DFL, \textit{ColNet} uses collaborative optimization so that nodes benefit from both task-coherent peers and cross-task exchanges. \figurename~\ref{fig:architecuture} outlines a five-stage recurrent workflow.

\begin{enumerate}
    \item {Local Learning:} Each node trains on its local dataset and updates both the shared \emph{backbone} and its \emph{task-specific layers}.
    \item {Model Exchanging:} Nodes exchange their current \emph{backbone} parameters with peers in the DFL overlay. Task-specific layers are not shared.
    \item {Adaptive Clustering and Intra-Group Aggregation:} Using model-sensitivity and data-sensitivity indicators derived from Stage 2, nodes are automatically grouped into task-coherent clusters. Within each cluster, nodes perform \emph{intra-group} aggregation on the backbone (FedAvg), while task-specific layers remain private to preserve task diversity. A cluster leader is designated to coordinate cross-group exchange.
    \item {Cross-Group Aggregation:} Cluster leaders exchange their group-averaged backbone. Each leader applies a simplified HCA aggregation to reduce conflicting gradients across tasks, producing an updated backbone that integrates complementary information from other groups.
    \item {Model Redistribution:} Leaders redistribute the updated backbone to all members of the corresponding clusters. Nodes synchronize their backbone parameters and resume Stage 1 with refreshed initialization.
\end{enumerate}

\begin{algorithm}[t]
\small
\caption{\textit{ColNet} Five-Stage Client Workflow (Node $i$)}
\label{alg:colnet_main}
\begin{algorithmic}[1]
\Require Graph $G=(\mathcal{V},\mathcal{E})$, neighbors $\mathcal{N}(i)$, data $\mathcal{D}_i$, loss $\ell$, rounds $R$, local epochs $E$, clusters $K$, weights $\alpha,\beta,\gamma$, temps $\tau_{\text{loss}},\tau_{J}$, leader set $\mathcal{L}_{\text{lead}}$
\State Initialize $\mathbf{w}_i=(B_i,H_i)$; $B_i^{\text{prev}}\gets B_i$
\For{$r=1$ to $R$}
  \State \textbf{Stage 1: Local Learning} \Comment{optimize $f_i$ on $\mathcal{D}_i$}
  \State Update $(B_i,H_i)$ by $E$-epoch local training with loss $\ell$
  \State \textbf{Stage 2: Model Exchanging} \Comment{backbone only}
  \State Exchange $B_i$ with all $j\in\mathcal{N}(i)$ to obtain $\{B_j\}_{j\in\mathcal{N}(i)}$
  \State \textbf{Stage 3: Adaptive Clustering \& Intra-Group Aggregation}
  \State $(g_i,\,\mathcal{K}(i)) \gets \Call{TaskAwareClustering}{B_i,H_i,\{B_j\},\mathcal{D}_i}$
  \State $B_i \gets \Call{FedAvg}{\{B_k\}_{k\in\mathcal{K}(i)}\cup\{B_i\}}$ \Comment{heads $H$ stay local}
  \State \textbf{Stage 4: Cross-Group Aggregation} \Comment{leaders only}
  \If{$i\in\mathcal{L}_{\text{lead}}$}
    \State $\Delta \bar B_i \gets B_i - B_i^{\text{prev}}$
    \State $B_i \gets \Call{LeaderHCA}{B_i,\Delta \bar B_i,\mathcal{L}_{\text{lead}}}$
  \EndIf
  \State \textbf{Stage 5: Redistribution}
  \If{$i\in\mathcal{L}_{\text{lead}}$}
    \State Broadcast updated $B_i$ to all $k\in\mathcal{K}(i)$
  \Else
    \State Receive $B_{\text{leader}}$ from group leader; $B_i \gets B_{\text{leader}}$
  \EndIf
  \State $B_i^{\text{prev}}\gets B_i$
\EndFor
\end{algorithmic}
\end{algorithm}

\subsection{Learning Process}
Compared to standard DFL methods, the main novelty of \textit{ColNet} during training lies in introducing \emph{adaptive clustering}, \emph{intra-task}, and \emph{cross-task} aggregation. The learning process in each client is presented in Algorithm~\ref{alg:colnet_main}.

\paragraph{\textbf{Local Learning}}
Client \(i\) solves the local objective on \(\mathcal{D}_i\) for \(E\) epochs (cf. Eq.~\eqref{eq:dfmt_objective}), updating both \(B_i\) and \(H_i\).
\smallskip

\paragraph{\textbf{Model Exchanging}}
Clients exchange \emph{backbone} parameters with neighbors in \(G\), i.e., each node broadcasts \(B_i\) and receives \(\{B_j\}_{j\in\mathcal{N}(i)}\). Task heads \(H\) are not shared.
\smallskip

\paragraph{\textbf{Adaptive Clustering and Intra-Group Aggregation}}
Given received backbones $\{B_j\}_{j\in\mathcal{N}(i)}$, client $i$ forms temporary models
$M_0=(B_i,H_i)$ and $M_j=(B_j,H_i)$ for all $j\in\mathcal{N}(i)$ so that all models share the \emph{same} task head $H_i$.
This aligns the evaluation to client $i$'s task and avoids confounding from heterogeneous heads. The adaptive clustering is presented in Algorithm~\ref{alg:task_aware_clustering}.

\smallskip
\noindent\textit{(1) Three signals on local data.}
On $\mathcal{D}_i$, client $i$ computes three pairwise signals among the index set $\mathcal{I}=\{0\}\cup\mathcal{N}(i)$:
\begin{align}
&\text{Cosine similarity:} &&
C[p,q] \;=\; \frac{\langle \theta(M_p),\, \theta(M_q)\rangle}{\|\theta(M_p)\|\,\|\theta(M_q)\|},
\label{eq:cos}
\\
&\text{Empirical loss:} &&
\Lambda[p,q] \;=\; \mathbb{E}_{(x,y)\sim\mathcal{D}_i}\big[\ell\big(f(x;M_q),y\big)\big],
\label{eq:loss}
\\
&\text{Avg. Jacobian norm:} &&
J[p,q] \;=\; \mathbb{E}_{x\sim\mathcal{D}_i}\!\left[\left\|\,\frac{\partial f(x;M_q)}{\partial z(x;M_q)}\,\right\|_F\right].
\label{eq:jac}
\end{align}
Here $\theta(M)$ stacks the (flattened) backbone parameters of $M$ (heads are identical and omitted),
$f$ is the task output (e.g., logits), and $z(\cdot;M_q)$ denotes the \emph{backbone feature} right before the head.
The Jacobian norm measures the model’s local sensitivity to backbone features, serving as a proxy for model stability. 

Together, these three signals capture both model similarity and data-dependent sensitivity on $\mathcal{D}_i$, providing a balanced view that is less biased than any single metric.

\smallskip
\noindent\textit{(2) Symmetrization and temperature scaling.}
$\Lambda$ and $J$ are asymmetric because they evaluate $M_q$ on $\mathcal{D}_i$, thus, they are symmetrized via:
\begin{equation}
\begin{aligned}
\Lambda \leftarrow \tfrac{1}{2}(\Lambda+\Lambda^\top) \\
J \leftarrow \tfrac{1}{2}(J+J^\top).
\end{aligned}
\label{eq:sym}
\end{equation}

Then, they are converted into similarities via temperatures $\tau_{\text{loss}},\tau_J>0$:
\begin{equation}
\begin{aligned}
\Lambda_{\text{sim}}[p,q] = \exp\!\big(-\Lambda[p,q]/\tau_{\text{loss}}\big) \\
J_{\text{sim}}[p,q] = \exp\!\big(-J[p,q]/\tau_J\big).
\end{aligned}
\label{eq:soft-sims}
\end{equation}
In practice, this work \emph{standardizes} each matrix before exponentiation (z-score per-matrix) to balance scale across signals.

\smallskip
\noindent\textit{(3) Affinity fusion and self-confidence.}
This work fuses the three channels into an affinity matrix:
\begin{equation}
\begin{aligned}
S \;=\; \alpha\,C \;+\; \beta\,\Lambda_{\text{sim}} \;+\; \gamma\,J_{\text{sim}} \\ 
\qquad \alpha,\beta,\gamma \ge 0 \; \& \; \alpha{+}\beta{+}\gamma=1,
\end{aligned}
\label{eq:affinity}
\end{equation}

\smallskip
\noindent\textit{(4) Spectral clustering.}
Apply normalized spectral clustering on the affinity matrix $S$ to obtain $K$ clusters. 
Let $g_i$ be the label assigned to index $0$ (client $i$), and define its in-group peers as 
$\mathcal{K}(i)=\{k\in\mathcal{N}(i)\cup\{i\}\mid g_k=g_i\}$.

\smallskip
\noindent\textit{(5) Intra-group aggregation (backbone only).}
Within $\mathcal{K}(i)$, clients average backbones while keeping heads local:
\begin{equation}
\bar{\mathbf{w}}^B \;=\; \frac{1}{|\mathcal{K}(i)|}\!\sum_{k\in\mathcal{K}(i)} \mathbf{w}_k^B,
\qquad
B_i \leftarrow \bar{\mathbf{w}}^B.
\label{eq:intra-agg}
\end{equation}
This produces a group-coherent backbone $B_i$ tailored to $i$'s task via $H_i$.

\begin{algorithm}[t]
\small
\caption{\textsc{TaskAwareClustering} at Node $i$}
\label{alg:task_aware_clustering}
\begin{algorithmic}[1]
\Require $B_i,H_i$, neighbor backbones $\{B_j\}_{j\in\mathcal{N}(i)}$, data $\mathcal{D}_i$, loss $\ell$, $\alpha,\beta,\gamma$, $\tau_{\text{loss}},\tau_{J}$, clusters $K$
\Ensure Cluster label $g_i$, peer set $\mathcal{K}(i)$
\State Build models $M_0\!\gets\!(B_i,H_i)$ and $M_j\!\gets\!(B_j,H_i)$ for $j\!\in\!\mathcal{N}(i)$
\State Let index set $\mathcal{I}\!\gets\!\{0\}\cup\mathcal{N}(i)$; init $C,\Lambda,J\in\mathbb{R}^{|\mathcal{I}|\times|\mathcal{I}|}$
\ForAll{$p,q\in\mathcal{I}$}
  \State $C[p,q]\gets \Call{CosSim}{M_p,M_q}$
  \State $\Lambda[p,q]\gets \Call{AvgLoss}{M_q;\mathcal{D}_i,\ell}$
  \State $J[p,q]\gets \Call{AvgJacobianNorm}{M_q;\mathcal{D}_i}$
\EndFor
\State $\Lambda\gets(\Lambda+\Lambda^\top)/2$, \ $J\gets(J+J^\top)/2$
\State $\Lambda_{\text{sim}}[p,q]\gets \exp(-\Lambda[p,q]/\tau_{\text{loss}})$
\State $J_{\text{sim}}[p,q]\gets \exp(-J[p,q]/\tau_{J})$
\State $S\gets \alpha C + \beta \Lambda_{\text{sim}} + \gamma J_{\text{sim}}$; set $S[p,p]\gets 1$
\State Apply spectral clustering on $S$ into $K$ clusters; obtain $g_i$
\State $\mathcal{K}(i)\gets \{k\in\mathcal{N}(i)\cup\{i\}\mid g_k=g_i\}$
\State \Return $(g_i,\,\mathcal{K}(i))$
\end{algorithmic}
\end{algorithm}

\noindent\textit{(6) Leader Selection.} 
Once task groups are established, each group appoints a leader on a rotating basis to balance coordination overhead. 
In round $r$, one member serves as leader and coordinates inter-group aggregation; after this step, leadership rotates (round-robin) to the next member in the group. The outgoing leader announces the handover to its group and to peer leaders so that subsequent cross-group exchanges use the updated routing.

\paragraph{\textbf{Cross-Group Aggregation}}
Cross-group aggregation is challenging due to potential conflicts between updates produced by heterogeneous tasks. Let \(\mathcal{L}_{\text{lead}}\) denote the set of current group leaders and \(G_\ell = |\mathcal{L}_{\text{lead}}|\) the number of groups. Each leader \(i\!\in\!\mathcal{L}_{\text{lead}}\) computes a backbone delta to its previous round,
\begin{equation}
\Delta \mathbf{w}_i^B \;=\; \bar{\mathbf{w}}_i^B - \mathbf{w}_{i,\text{prev}}^B,
\end{equation}
exchanges \(\Delta \mathbf{w}_i^B\) with other leaders, and forms the mean update
\(
\Delta \bar{\mathbf{w}}^B = \tfrac{1}{G_\ell}\sum_{\ell\in\mathcal{L}_{\text{lead}}}\Delta \mathbf{w}_\ell^B.
\)

Inspired by FedHCA\(^2\)~\cite{lu2024fedhca2}, \textit{ColNet} adapts a gradient-alignment objective that seeks an aggregated update \(\tilde{U}\) close to \(\Delta \bar{\mathbf{w}}^B\) while improving its alignment with the worst-case task direction:
\begin{equation}
\label{eq:hca-optimization}
\begin{aligned}
& \max_{\tilde{U}} \min_i \langle \Delta \mathbf{w}_i^B, \tilde{U} \rangle 
\\ & \text{s.t.}\quad 
\|\tilde{U} - \Delta \bar{\mathbf{w}}^B\| \;\le\; c \,\|\Delta \bar{\mathbf{w}}^B\|
\end{aligned}
\end{equation}

where \(c\in[0,1)\) bounds the deviation from the mean update. A Lagrangian treatment yields the closed-form
\begin{equation}
\label{inter-task-aggregation}
\begin{aligned}
\tilde{U} \;=\; \Delta \bar{\mathbf{w}}^B \;+\; \frac{\sqrt{\phi}}{\|U_w\|}\,U_w\\
U_w \;=\; \frac{1}{G_\ell}\sum_{\ell\in\mathcal{L}_{\text{lead}}} w_\ell\,\Delta \mathbf{w}_\ell^B\\
\phi \!=\! c^2 \|\Delta \bar{\mathbf{w}}^B\|^2
\end{aligned}
\end{equation}
with nonnegative weights \(w_\ell\) determined by the chosen conflict metric. This update balances proximity to the mean and reduction of cross-task conflicts, as shown in Algorithm~\ref{alg:leader_hca}. Each leader applies \(\tilde{U}\) to its backbone and proceeds to redistribution.

\begin{algorithm}[t]
\small
\caption{\textsc{LeaderHCA} (Cross-Group Agg. at Leader $i$)}
\label{alg:leader_hca}
\begin{algorithmic}[1]
\Require Current backbone $B_i$, local delta $\Delta \bar B_i$, leader set $\mathcal{L}_{\text{lead}}$
\Ensure Updated backbone $B_i$
\ForAll{$\ell\in\mathcal{L}_{\text{lead}}\setminus\{i\}$}
  \State Send $\Delta \bar B_i$ to $\ell$;\ Receive $\Delta \bar B_\ell$ from $\ell$
\EndFor
\State $\tilde{U}_i \gets \Call{HCA}{\Delta \bar B_i,\{\Delta \bar B_\ell\}_{\ell\in\mathcal{L}_{\text{lead}}\setminus\{i\}}}$ \Comment{see Eq.\,(\ref{inter-task-aggregation})}
\State $B_i \gets B_i + \tilde{U}_i$
\State \Return $B_i$
\end{algorithmic}
\end{algorithm}

\paragraph{\textbf{Model Redistribution}}
Group leaders broadcast the updated backbone to their members, who synchronize \(B\) and continue with the next round, while task heads remain private. The rotating-leader policy is then applied to select successors for the following cross-task aggregation.

Stages (2)–(3) adapt collaboration to both model- and data-driven signals, which improves stability under label and task heterogeneity. Stage (4) integrates complementary cross-task information while mitigating gradient conflicts. The five-stage loop repeats until a stopping criterion is met.

\section{Deployment and Evaluation}
\label{sec:experiments}
This section details the deployment and experimental evaluation of \textit{ColNet}, covering the targeted multi-task learning scenarios, the chosen evaluation metrics, and the resulting performance outcomes \footnote{Code available at: https://github.com/Cyber-Tracer/asfdfmtl}.

\subsection{Scenarios, Datasets, and Models}
Two heterogeneity scenarios are considered for evaluation: \emph{label heterogeneity} and \emph{task heterogeneity}.

\textbf{Label heterogeneity} arises when clients train on different subsets of class labels, which requires splitting the model into a shared backbone and task-specific layers to protect privacy and maintain performance. Clients with the same labels form a single task group, sharing only the backbone parameters. The \textbf{CIFAR-10} dataset \cite{krizhevsky2009learning} was selected to evaluate label heterogeneity, where each image belongs to one of ten classes. To simulate label heterogeneity, the dataset is partitioned into two super-classes: \textbf{animals} (bird, cat, deer, dog, frog, horse) and \textbf{objects} (airplane, automobile, ship, truck). Within each super-class, samples are split evenly and assigned to three clients in an IID manner.

The dataset is split into 50,000 images for training and 10,000 for testing, further subdivided into animal or object sets. Each client’s training subset is then partitioned (80\% train and 20\% validation), and all clients use the same test subset in a task group. Data augmentation techniques, such as random cropping, flipping, and color jittering are applied to mitigate overfitting.

All clients train a ResNet-18 \cite{he2016deep} backbone and separate task layers to accommodate different output dimensions (four classes for objects, six for animals). A Cross-Entropy loss and SGD optimizer (learning rate 0.01, momentum 0.9, weight decay 1e-3) are employed.

\textbf{Task heterogeneity} is more complex, as clients may work on entirely different tasks (e.g., image classification and object detection), each requiring distinct final layers, loss functions, and optimizers. Dividing the model into a shared backbone and task-specific layer can facilitate aggregation in this scenario. The \textbf{CelebA} dataset \cite{liu2015faceattributes} is used to evaluate task heterogeneity. It contains over 200,000 face images with up to 40 binary attributes and five landmark annotations, making it suitable for multiple tasks, including face attribute recognition and landmark detection \cite{woubie2024maintaining,rasouli2020fedgan,mortaheb2022fedgradnorm}.

To simulate task heterogeneity, two distinct task groups are defined: \textbf{multi-label face attribute classification} and \textbf{face landmark detection}. A subset comprising 80\% of the dataset is used for training/validation, created by first selecting 40\% for each task group and assigning it evenly across that group’s clients; the remaining 20\% serves as a common test set. Each client then partitions its local portion into 80\% training and 20\% validation. Standard augmentations (random rotation and horizontal flips) are applied to mitigate overfitting.

Both tasks employ ResNet-18 backbone and distinct task layers: (a) \textbf{Face attribute classification} uses Binary Cross Entropy loss and an AdamW optimizer 5e-5 learning rate, 5e-3 weight decay; (b) \textbf{Face landmark detection} uses Mean Squared Error loss and an Adam optimizer 5e-4 learning rate. 

In the experiments, the overlay network uses a fully connected topology, and nodes have no knowledge of other nodes’ task types. After each local training round, each node disseminates only its backbone parameters to all other nodes in the network. Adaptive task clustering is then performed based on the received backbones and local data. Normalized spectral clustering is used with the number of clusters set to $k$=2, after which intra-group aggregation and subsequent steps proceed.

The experiments report four evaluation metrics: loss, precision, recall, and F1-Score. Since face landmark detection is a regression task, only the loss metric is used to evaluate this subtask. To handle class imbalance and multi-class scenarios, \textit{micro} averaging strategy is used, aggregating metrics across all classes within each task group. Metrics are kept separate for different groups because their labels (in the label heterogeneity scenario) or tasks (in the task heterogeneity scenario) are distinct. 

\subsection{Federation Setup}
For the federated setup, six nodes are split into two task groups, each fully connected internally, with inter-group communication handled by leader nodes. Fifteen training rounds are performed, each comprising two local epochs followed by intra-group and cross-group model aggregation. 
To investigate aggregation frequency, the experiment also compares 2, 3, and 5 local epochs per round.

Three baselines are implemented to assess the proposed \textit{ColNet} framework:
\begin{enumerate}
\item No aggregation: Nodes train only on local data.  
\item Intra-aggregation only: Nodes aggregate their backbone parameters within each group, with no cross-group exchange.  
\item FedPre (intra- and cross-aggregation): Nodes use FedPre~\cite{arivazhagan2019federated} for both intra-group and cross-group backbone aggregation.  
\end{enumerate}
By contrast, \textit{ColNet} combines FedPre for intra-group aggregation with HCA-based cross-group aggregation, enabling more effective handling of diverse tasks.

\subsection{Impact of Aggregation Frequency}
To maximize multi-task learning performance, it is important to determine the optimal aggregation frequency, \textit{i.e.,} how many local epochs to run per training round. Hence, the initial experiment compares \textit{ColNet}’s performance with 2, 3, and 5 local epochs under two evaluation scenarios, thereby identifying the most suitable hyperparameter.

\begin{table}[h]
\caption{Average Client Test Metrics of \textit{ColNet} for Two Evaluation Scenarios with Different Local Epoch Setups}
\label{tab:local_epoch}
\resizebox{\columnwidth}{!}{%
\begin{tabular}{cccllll}
\toprule
\multicolumn{1}{l}{\textbf{Scenario}} & \multicolumn{1}{l}{\textbf{Subtask}} & \textbf{Epochs} & \textbf{Loss} & \textbf{Precision}  & \textbf{Recall}     & \textbf{F1-Score}   \\ \hline
\multirow{6}{*}{\begin{tabular}[c]{@{}c@{}}Label\\  Heterogeneity\\ (CIFAR-10)\end{tabular}}  & \multirow{3}{*}{\begin{tabular}[c]{@{}c@{}}Animal \\  Classification\end{tabular}}     & 2  & \textbf{0.673} & \textbf{0.779} & \textbf{0.768} & \textbf{0.769} \\
 &   & 3 & 0.786  & 0.764 & 0.732 & 0.729 \\
 &   & 5 & 0.801  & 0.743 & 0.727 & 0.717 \\ \cmidrule{2-7}
 & \multirow{3}{*}{\begin{tabular}[c]{@{}c@{}}Object\\  Classification\end{tabular}}   & 2 & 0.370 & 0.888 & 0.878 & 0.877 \\
 &    & 3 & \textbf{0.333} & \textbf{0.895} & \textbf{0.891} & \textbf{0.891} \\
 &   & 5 & 0.383  & 0.884 & 0.874 & 0.874 \\ \hline
 \multirow{6}{*}{\begin{tabular}[c]{@{}c@{}}Task\\  Heterogeneity\\ (CelebA)\end{tabular}} & \multirow{3}{*}{\begin{tabular}[c]{@{}c@{}}Attribute\\  Classification\end{tabular}}  & 2  & \textbf{0.231} & \textbf{0.741} & \textbf{0.545} & \textbf{0.605} \\
 &      & 3 & 0.233  & 0.735 & 0.542 & 0.598 \\
 &   & 5 & 0.237  & 0.733 & 0.522 & 0.579 \\  \cmidrule{2-7}
 & \multirow{3}{*}{\begin{tabular}[c]{@{}c@{}}Landmark\\  Detection\end{tabular}} & 2 & 5.299  & - & - & - \\
 &  & 3 & \textbf{5.224} & - & - & - \\
 &   & 5 & 5.762  & - & - & -    \\ \bottomrule
\end{tabular}%
}
\end{table}

\begin{figure}[h]
    \centering
    \includegraphics[width=1\columnwidth]{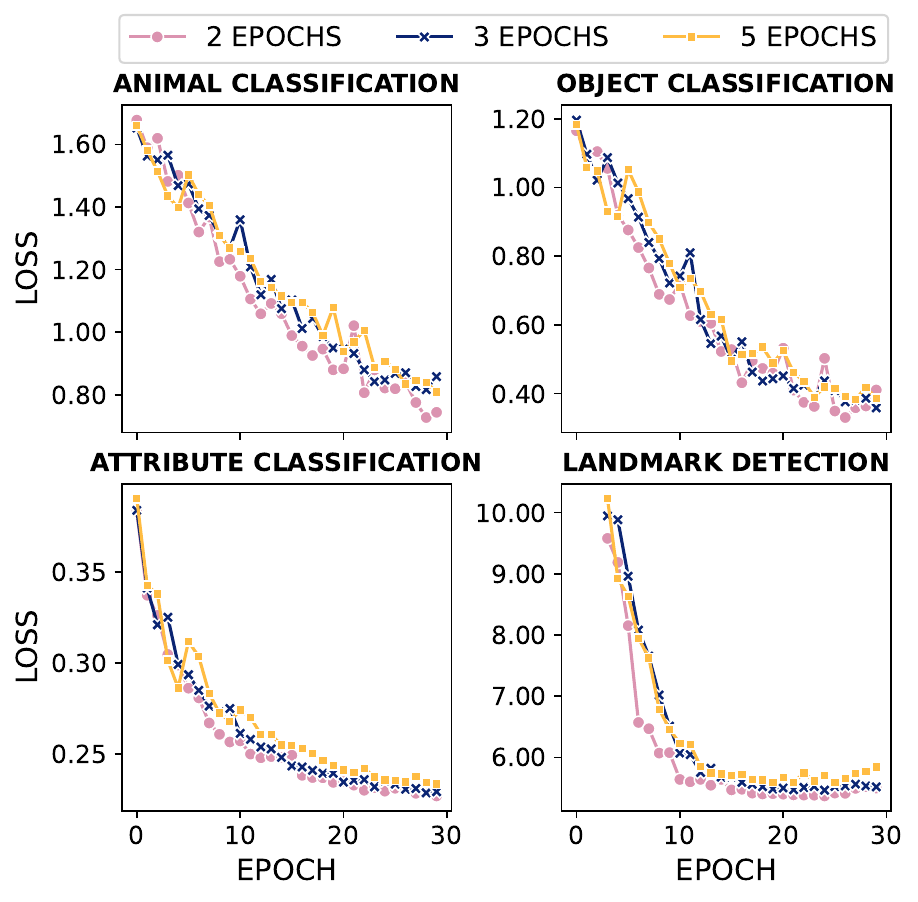}
    \caption{Average Client Validation Loss in Each Epoch with Different Local Epoch Setups}
    \label{fig:epochs}
\end{figure}

\tablename~\ref{tab:local_epoch} presents the results for four subtasks: animal classification, object classification, face attribute classification, and face landmark detection. Since face landmark detection is a regression task, only the loss metric is reported for that subtask. Besides, \figurename~\ref{fig:epochs} presents the epoch-wise validation loss for each subtask with different local training epoch settings.

In the label heterogeneity scenario (\textit{i.e.,} CIFAR-10), using two epochs per aggregation achieves the best performance for the animal group, while the object group slightly trails the three-epoch scheme. However, the difference is negligible, and both precision and recall remain closely aligned. From a training perspective (see Figure~\ref{fig:epochs}), the two-epoch setup converges more smoothly and exhibits lower loss. Despite the minor drop for the object group, the overall benefits of two-epoch intervals prevail.

Frequent aggregations benefit face attribute classification in the task heterogeneity scenario (\textit{i.e.,} CelebA). While the landmark detection task achieves its best results with three epochs, the two-epoch setting converges more quickly and smoothly. Consequently, two local training epochs per round are adopted in all remaining experiments.

\subsection{Impact of the Privatization Degree}
When deploying the backbone and task layers, a crucial hyperparameter concerns how to allocate these two parts of the network architecture. Although both scenarios in this work rely on ResNet-18 as the base model, determining which layers form the backbone and which form the task layers still requires careful consideration. Retaining more of the original, lower-level ResNet-18 layers in the backbone increases the amount of shared information among nodes, which may be detrimental for highly heterogeneous tasks. Conversely, expanding the upper residual layers into the task layer offers a higher degree of privatization but may weaken the effectiveness of aggregations.

\begin{table}[b]
\caption{Average Client Test Metrics of \textit{ColNet} for Two Evaluation Scenarios with Different Privatization Setups}
\label{tab:task_layer}
\resizebox{\columnwidth}{!}{%
\begin{tabular}{cccllll}
\toprule
\multicolumn{1}{l}{\textbf{Scenario}} & \multicolumn{1}{l}{\textbf{Subtask}} & \textbf{Privatization} & \textbf{Loss} & \textbf{Precision} & \textbf{Recall} & \textbf{F1-Score}\\ \hline
\multirow{6}{*}{\begin{tabular}[c]{@{}c@{}}Label\\      Heterogeneity\\ (CIFAR-10)\end{tabular}} & \multirow{3}{*}{\begin{tabular}[c]{@{}c@{}}Animal \\      Classification\end{tabular}} & BL-0 & 0.706 & \textbf{0.786} & 0.759 & 0.758 \\
 &  & BL-1 & \textbf{0.673} & 0.779 & \textbf{0.768} & \textbf{0.769} \\
 &  & BL-2 & 0.810 & 0.739 & 0.720 & 0.716 \\ \cmidrule{2-7}
 & \multirow{3}{*}{\begin{tabular}[c]{@{}c@{}}Object\\      Classification\end{tabular}} & BL-0 & \textbf{0.313} & \textbf{0.900} & \textbf{0.895} & \textbf{0.895} \\
 &  & BL-1 & 0.370 & 0.888 & 0.878 & 0.877 \\
 &  & BL-2 & 0.371 & 0.881 & 0.876 & 0.875 \\ \hline
\multirow{6}{*}{\begin{tabular}[c]{@{}c@{}}Task\\      Heterogeneity\\ (CelebA)\end{tabular}} & \multirow{3}{*}{\begin{tabular}[c]{@{}c@{}}Attribute\\      Classification\end{tabular}} & BL-0 & \textbf{0.230} & 0.738 & 0.541 & 0.599 \\
 &  & BL-1 & 0.231 & \textbf{0.741} & \textbf{0.545} & \textbf{0.605} \\
 &  & BL-2 & 0.236 & 0.734 & 0.533 & 0.592 \\ \cmidrule{2-7}
 & \multirow{3}{*}{\begin{tabular}[c]{@{}c@{}}Landmark\\      Detection\end{tabular}} & BL-0 & \textbf{5.285} & - & - & - \\
 &  & BL-1 & 5.299 & - & - & - \\
 &  & BL-2 & 5.829 & - & - & -  \\ \bottomrule
\end{tabular}%
}
\end{table}

\begin{figure}[h]
    \centering
    \includegraphics[width=1\columnwidth]{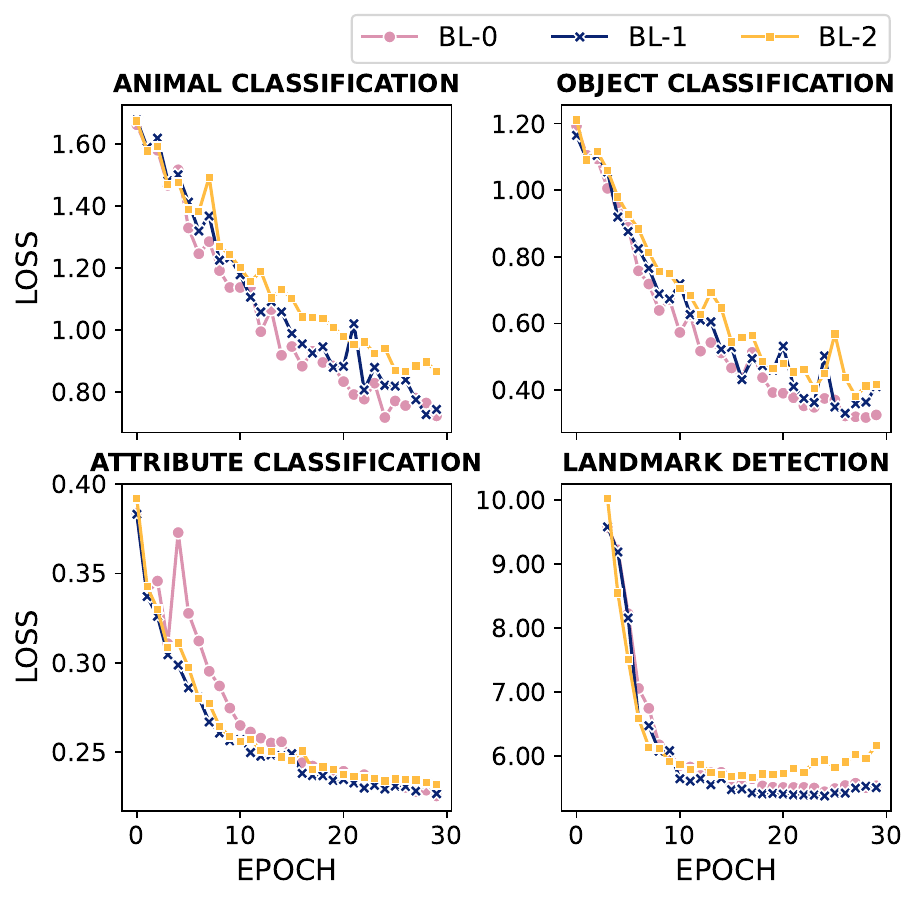}
    \caption{Average Client Validation Loss in Each Epoch with Different Privatization Setups}
    \label{fig:bls}
\end{figure}

To evaluate how different levels of privatization impact multi-task learning, three configurations are compared: (i) no residual layers are privatized (BL-0), (ii) the last residual layer is part of the task layer (BL-1), and (iii) the last two residual layers are part of the task layer (BL-2). BL-0 represents the lowest degree of privatization, whereas BL-2 represents a high degree. 

\tablename~\ref{tab:task_layer} compares the performance of \textit{ColNet} client models under these three settings in both scenarios. Besides, \figurename~\ref{fig:bls} presents the validation loss in each epoch during the training process. Overall, BL-2 shows the weakest performance, implying that retaining more residual layers in the backbone benefits DFMTL by providing sufficient neurons to exchange knowledge across different tasks. At the same time, a suitable number of task layers enables proper personalization under varied conditions. Drawing on the results from both scenarios, a moderate privatization strategy—treating the last residual block and the output layer of ResNet-18 as task layers (\textit{i.e.,} BL-1) proves optimal and is therefore adopted for subsequent experiments.

\subsection{Comparison Among Aggregation Schemes}
Having established an appropriate aggregation frequency and layer privatization strategy, the \textit{ColNet} framework is now compared with other aggregation schemes. The main objective is to determine whether clients that leverage cross-task knowledge achieve better performance on their own tasks.

This experiment compares the performance of \textit{ColNet} with (i) No aggregation (NO AGG.), (ii) Intra-aggregation (INTRA-AGG.), and (iii) FedPre with intra- and cross-aggregation (FEDPRE I\&C), in label heterogeneity and task heterogeneity scenarios. \tablename~\ref{tab:agg} presents the test metrics of different aggregation schemes in these two scenarios after 15 federated rounds with a total of 30 epochs.

\begin{table}[h!]
\caption{Average Client Test Metrics for Different Aggregation Schemes}
\label{tab:agg}
\resizebox{\columnwidth}{!}{%
\begin{tabular}{cccllll}
\toprule
\textbf{Scenario} & \textbf{Subtask} & \textbf{Aggregation} & \textbf{Loss} & \textbf{Precision} & \textbf{Recall} & \textbf{F1-Score} \\ 
\midrule
\multirow{8}{*}{\begin{tabular}[c]{@{}c@{}}Label\\ Heterogeneity\\ (CIFAR-10)\end{tabular}} 
 & \multirow{4}{*}{\begin{tabular}[c]{@{}c@{}}Animal\\ Classification\end{tabular}} 
   & NO AGG.& 0.938 & 0.699 & 0.675 & 0.669 \\
 &  & INTRA-AGG.& 0.953 & 0.723 & 0.680 & 0.679 \\
 &  & FEDPRE(I\&C) & 0.853 & 0.730 & 0.700 & 0.689 \\
 &  & \textbf{ColNet}  & \textbf{0.673} & \textbf{0.779} & \textbf{0.768} & \textbf{0.769} \\
\cmidrule(lr){2-7}
 & \multirow{4}{*}{\begin{tabular}[c]{@{}c@{}}Object\\ Classification\end{tabular}} 
   & NO AGG.& 0.464 & 0.846 & 0.840 & 0.840 \\
 &  & INTRA-AGG.& 0.409 & 0.867 & 0.855 & 0.856 \\
 &  & FEDPRE(I\&C) & \textbf{0.346} & 0.879 & 0.875 & 0.875 \\
 &  & \textbf{ColNet}  & 0.370 & \textbf{0.888} & \textbf{0.878} & \textbf{0.877} \\
\midrule
\multirow{8}{*}{\begin{tabular}[c]{@{}c@{}}Task\\ Heterogeneity\\ (CelebA)\end{tabular}} 
 & \multirow{4}{*}{\begin{tabular}[c]{@{}c@{}}Attribute\\ Classification\end{tabular}} 
   & NO AGG.& 0.244 & 0.722 & 0.511 & 0.570 \\
 &  & INTRA-AGG.& 0.236 & 0.733 & 0.526 & 0.585 \\
 &  & FEDPRE(I\&C) & 0.239 & 0.726 & 0.513 & 0.571 \\
 &  & \textbf{ColNet}  & \textbf{0.231} & \textbf{0.741} & \textbf{0.545} & \textbf{0.605} \\
\cmidrule(lr){2-7}
 & \multirow{4}{*}{\begin{tabular}[c]{@{}c@{}}Landmark\\ Detection\end{tabular}} 
   & NO AGG.& 6.054 & - & - & - \\
 &  & INTRA-AGG.& 5.392 & - & - & - \\
 &  & FEDPRE(I\&C) & 5.325 & - & - & - \\
 &  & \textbf{ColNet}  & \textbf{5.299} & - & - & - \\ \bottomrule
\end{tabular}%
}
\scriptsize
\centering
[NO AGG.: No aggregation; INTRA-AGG.: Intra-aggregation; \\ FEDPRE(I\&C): FedPre with intra- and cross-aggregation]
\end{table}

\tablename~\ref{tab:agg} compares \textit{ColNet} with three reference aggregation schemes under both the label heterogeneity and task heterogeneity scenarios. The results demonstrate that \textit{ColNet} consistently outperforms the other methods across multiple datasets in multi-task settings. 

In the label heterogeneity scenario, \textit{ColNet} achieves the highest performance in both the animal and object classification subtasks. Notably, the F1-Score for animal classification exceeds the reference aggregation schemes by more than 0.08, signifying a substantial improvement. The findings further highlight that collaborative learning in a federated environment produces clearly superior results compared to no aggregation at all. Aggregating only within task groups increases precision by about 0.02, while employing an intra- and inter-group FedPer approach improves performance even more. Besides, \textit{ColNet}’s cross-group HCA method effectively resolves conflicting gradients among heterogeneous tasks, allowing nodes to absorb knowledge from tasks beyond their own and enhancing local task performance.

A similar pattern is presented in the task heterogeneity scenario. Models without aggregation perform worst, whereas introducing inter-group aggregation yields a 0.01 gain in precision and a notable reduction in loss for the landmark detection subtask. With \textit{ColNet}, both subtasks attain the best results: the F1-Score rises by 0.03, precision increases by 0.02, and the landmark detection loss drops by 0.8. These outcomes underscore \textit{ColNet}’s clear advantages over other aggregation strategies in DFMTL.

\begin{figure}[h]
    \centering
    \includegraphics[width=1\columnwidth]{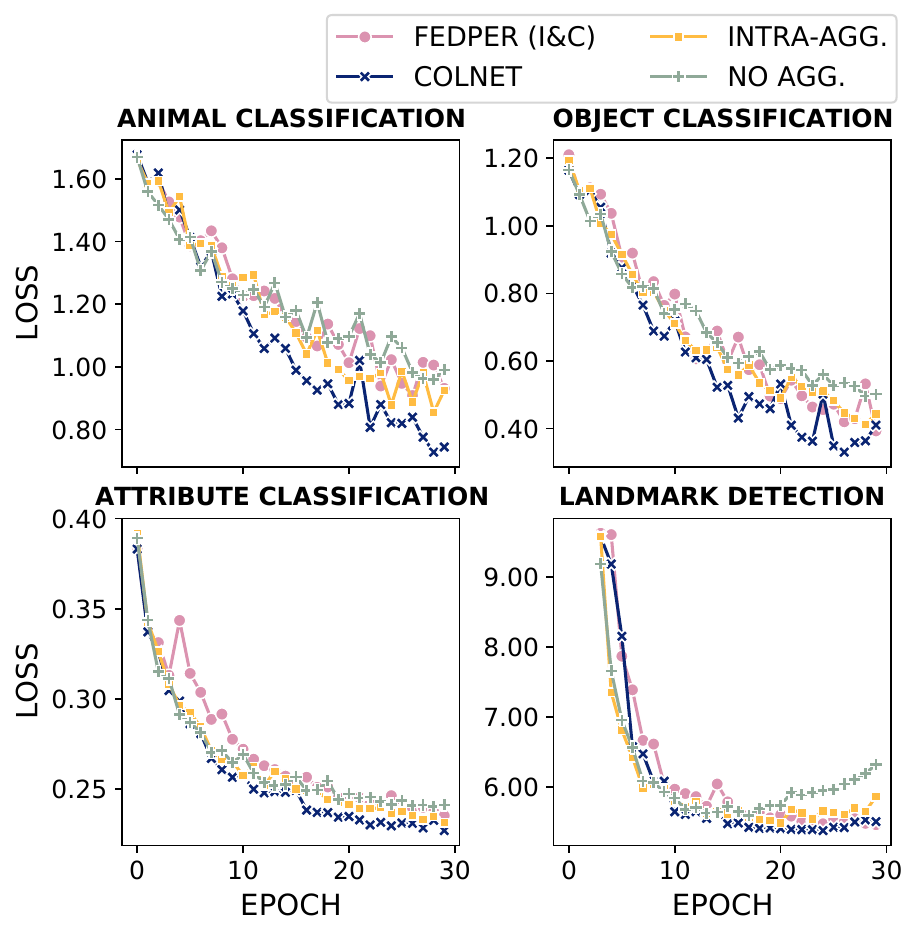}
    \caption{Average Client Validation Loss in Each Epoch for Different Aggregation Schemes}
    \label{fig:val_loss}
\end{figure}

\figurename~\ref{fig:val_loss} presents the average client validation loss across epochs with using different aggregation schemes. Across all tasks, \textit{ColNet} consistently achieves the lowest validation loss, highlighting its effectiveness in reducing errors during training. This advantage is particularly pronounced in animal classification and landmark detection, where \textit{ColNet}’s smoother and more stable convergence clearly outperforms the other methods. The results underscore \textit{ColNet}’s ability to manage both inter-task and intra-task heterogeneity more effectively than simpler aggregation schemes.

In contrast, FedPre and intra-aggregation show higher loss values and less stable convergence trends, while no aggregation performs the worst in all scenarios, reinforcing the critical importance of aggregation in federated multi-task learning. Notably, \textit{ColNet} demonstrates a significant advantage by mitigating gradient conflicts and enabling knowledge sharing. These findings affirm \textit{ColNet}’s superiority in DFMTL environments.

\subsection{Robustness Analysis of \textit{ColNet}}
In the preceding experiments, all participating clients were assumed benign (honest and non-adversarial). In decentralized learning, this assumption often breaks down: some clients may behave maliciously, most notably via poisoning attacks that corrupt model updates or local data, degrade performance, and undermine model usability~\cite{feng2024dart}. Accordingly, this study evaluates \textit{ColNet}’s resilience to multiple poisoning settings, covering both model-poisoning and data-poisoning variants, and analyzes its model robustness.

\subsubsection{Poisoning Attack Strategies}
This implements four poisoning attack strategies to evaluate the model robustness of the proposed \textit{ColNet}, covering data poisoning attacks (untargeted label flipping), model poisoning attacks (scaled boost and AT2FL~\cite{Sun2022DataPoisoning}), and aggregation manipulation.

\textbf{Untargeted Label Flipping.} Before local training, labels are corrupted to inject label noise: for multi-label tasks, each label is flipped with probability $p$ (Bernoulli mask), and for single-label tasks, the true class is replaced with a uniformly sampled different class with probability $p$. The operation degrades overall generalization without steering predictions toward a specific alternative.

\textbf{Scaled Boost.} After local training but before upload, the client scales its update by a factor $s>1$ and uploads $s\cdot\Delta_i$, optionally preserving non-trainable statistics and applying clipping or BN re-estimation to reduce detectability, thereby amplifying the client’s influence on global updates and inducing overshoot or instability.

\textbf{AT2FL (Inner-Loop Adversarial Data Poisoning).} During local training, for a small number of mini-batches the attacker computes input gradients and crafts adversarial inputs $\mathbf{x}_{\text{adv}}=\mathrm{Proj}_{\mathcal{X}}(\mathbf{x}+\epsilon\,\mathrm{sign}(\nabla_{\mathbf{x}}L))$, performs additional optimization steps on these perturbed samples to steer local parameters toward directions that increase future loss on target nodes, then uploads the manipulated backbone.

\textbf{Aggregation Manipulation (Malicious Aggregation Filter).} At aggregation, the aggregator replaces or perturbs the conflict-averse aggregate $\Delta_{\text{HCA}}=\mathcal{A}_{\text{HCA}}(\{\Delta_i\})$ with $\tilde{\Delta}_{\text{agg}}=\mathcal{F}_{\text{mal}}(\Delta_{\text{HCA}},\{\Delta_i\})$, thereby compromising cross-task reconciliation or implanting group-level backdoors.

\subsubsection{Attack evaluation setup}

The robustness analysis experiments follow the same base training and evaluation protocol as the model-performance experiments. The only change is that, in each attack trial, a single client is designated as the \emph{attacker} while all other clients remain benign. Comparative runs are executed in two modes: (a) \emph{clean baseline} (no attacker present) and (b) \emph{attacked} (one attacker active). Robustness is measured by the impact of the attack on the performance of benign clients; that is, if the loss and F1-Score of benign clients degrade only mildly under attack relative to the clean baseline, the system is considered more robust.

The dataset-specific attacker placement used in the experiments is as follows:
(i) \textbf{CIFAR-10 (label-heterogeneous setting).} The first node in the Animal Classification (AN\_N0) task is set as the attacker; the remaining nodes are benign. \\ 
(ii) \textbf{CelebA (task-heterogeneous setting).} The first node in the Attribute Classification (AC\_N0) task is set as the attacker; the remaining nodes are benign.

\subsubsection{Robustness Analysis Results}
\begin{figure}[h]
    \centering
    \includegraphics[width=1\columnwidth]{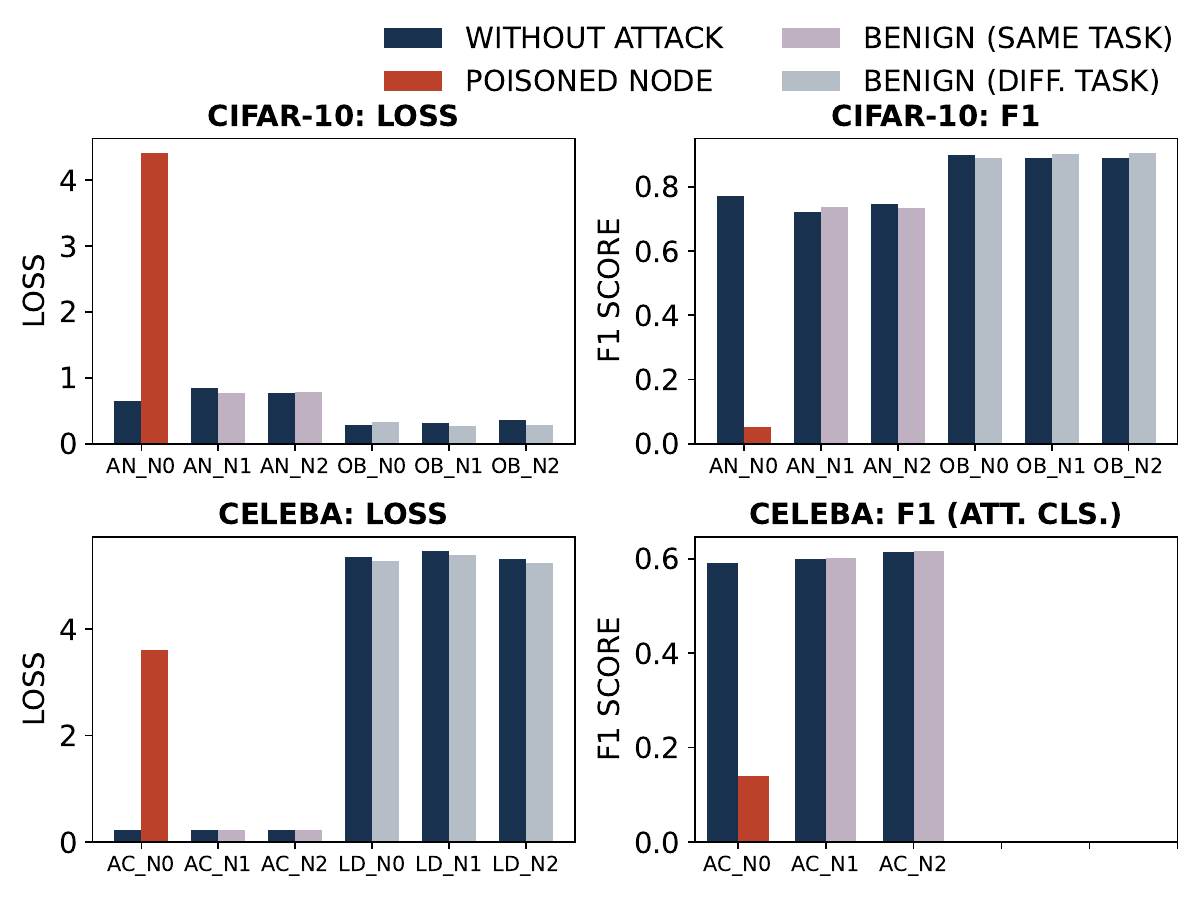}
    \caption{Impact of Untargeted Label Flipping Attack on CIFAR-10 and CelebA Datasets}
    \label{fig:untargeted_LF}
\end{figure}

\paragraph{Untargeted label flipping.}
This attack flips labels in the attacker’s local dataset before training. As shown in \figurename~\ref{fig:untargeted_LF}, its effect is highly concentrated on the poisoned node: e.g., on CIFAR-10 the poisoned AN\_N0’s loss increased from $\approx0.65$ to $\approx4.41$ and its F1 dropped from $0.77$ to $0.052$ (a $\sim0.72$ absolute F1 loss). By contrast, same-group mates  AN\_N1/N2) experienced negligible aggregate harm (mean F1 stayed roughly the same: $\sim0.734$ $\to$ $\sim0.736$), and cross-group object nodes (OB\_N\*) were essentially unaffected. A similar pattern holds on CelebA dataset: the poisoned attribute node (AC\_N0) suffered a large F1 drop (0.591 $\to$ 0.139), while the other attribute nodes showed small or no degradation.

\begin{figure}[b]
    \centering
    \includegraphics[width=1\columnwidth]{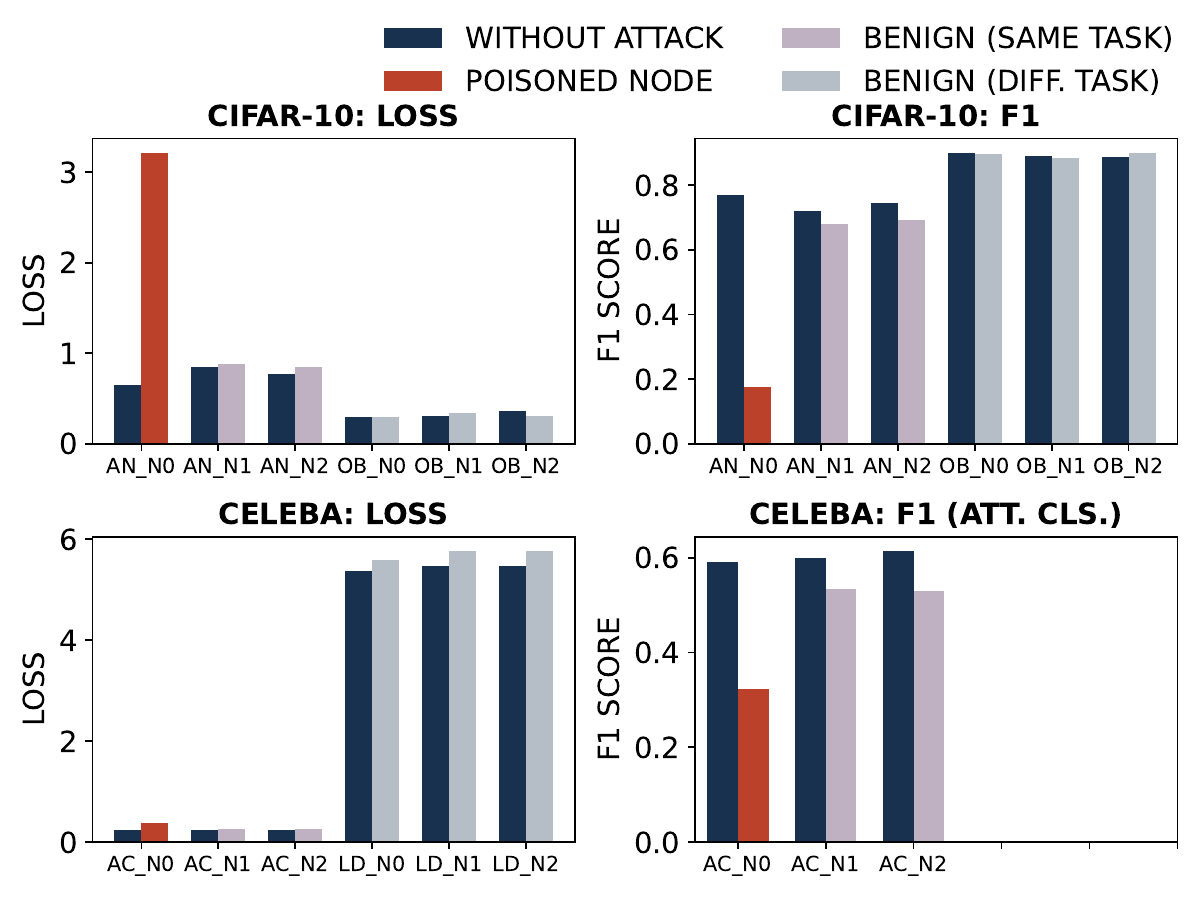}
    \caption{Impact of Scaled Boost Attack on CIFAR-10 and CelebA Datasets}
    \label{fig:Scaled_Boost}
\end{figure}

\paragraph{Scaled Boost (model poisoning).}
Scaled Boost amplifies a client’s update before upload (here \(\times5\) amplified). As shown in \figurename~\ref{fig:Scaled_Boost}, the poisoned client incurs the largest absolute degradation, for CIFAR-10 AN\_N0 F1 fell from $0.77$ to $\approx0.18$, but unlike simple label flips there is measurable collateral damage to same-group benign clients: AN\_N1/N2 saw modest F1 drops (each $\approx 0.04\mbox{--}0.05$ absolute), while cross-group object nodes remained effectively stable (changes $\approx$ 0 or within noise). On CelebA, the attacker’s attribute node (AC\_N0) lost $\sim\!0.27$ F1, and the other attribute nodes suffered smaller but non-negligible drops. 

\begin{figure}[t]
    \centering
    \includegraphics[width=1\columnwidth]{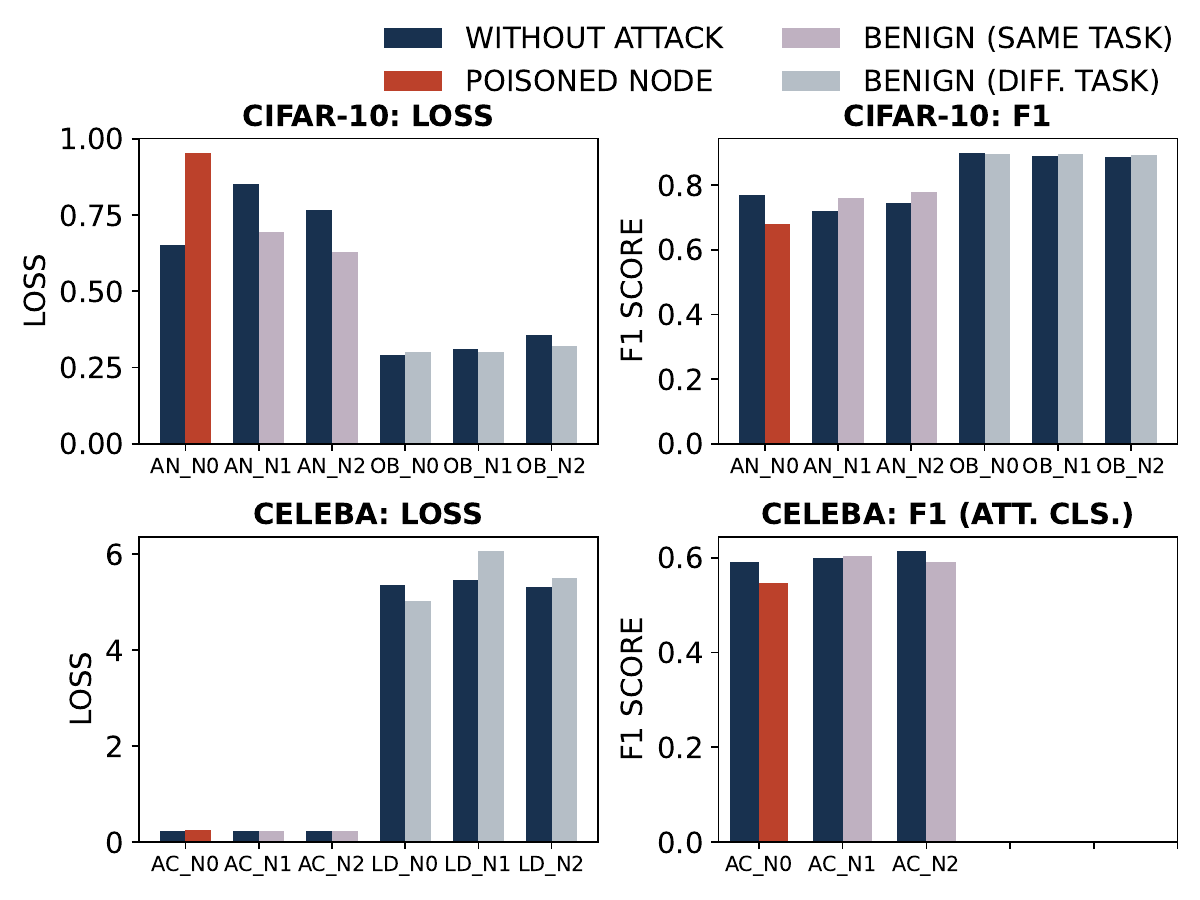}
    \caption{Impact of AT2FL Attack on CIFAR-10 and CelebA Datasets}
    \label{fig:at2fl}
\end{figure}

\paragraph{AT2FL (inner-loop adversarial poisoning).}
AT2FL crafts adversarial inputs during a few local mini-batches to bias the local update towards directions that maximize future loss on target nodes. As shown in \figurename~\ref{fig:at2fl}, for CIFAR-10, AN\_N0’s F1 decreased from $0.77$ to $\approx0.68$, while same-group mates (AN\_N1/N2) in some runs even improved slightly, indicating that adversarially nudging shared representations can have non-uniform, task-dependent side effects. Cross-group object nodes remained essentially stable (very small changes). On CelebA the AT2FL perturbations caused small-to-moderate F1 decreases on the attribute nodes (e.g., AC\_N0: $0.591\to0.547$).

\begin{figure}[b]
    \centering
    \includegraphics[width=1\columnwidth]{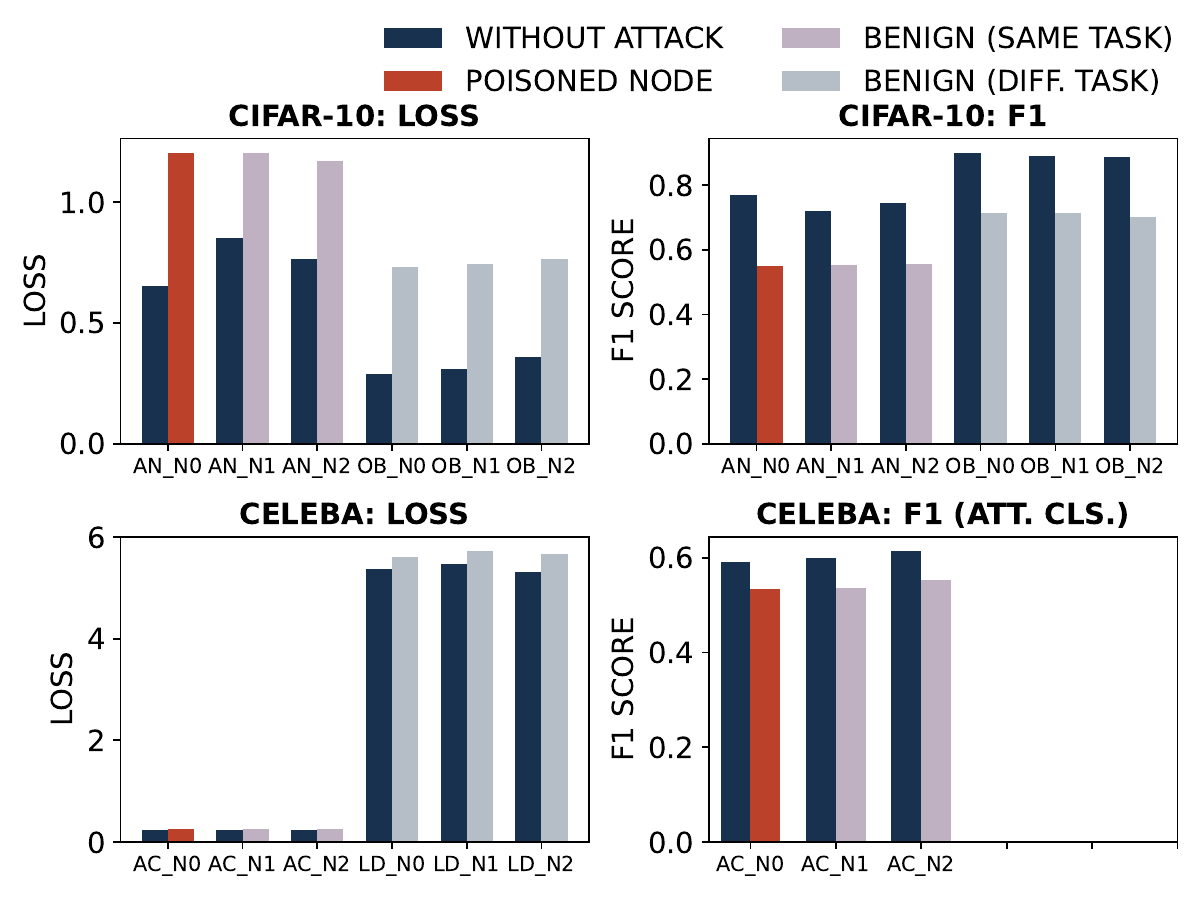}
    \caption{Impact of Aggregation manipulation Attack on CIFAR-10 and CelebA Datasets}
    \label{fig:hca}
\end{figure}

\paragraph{Aggregation manipulation.}
Tampering with the aggregation routine (replacing or perturbing the conflict-averse solution) is the most system-wide attack. As shown in \figurename\ref{fig:hca}, in CIFAR-10 the malicious aggregation caused sizeable F1 drops across both same-group and cross-group nodes: AN nodes’ F1 decreased from $\sim0.77$ to $\sim0.55$ (same-group mean drop $\approx0.18$), and object nodes’ F1 also fell substantially (cross-group mean drop $\approx0.18$). On CelebA the aggregation attack produced roughly uniform declines across attribute nodes (each AC\_N\* lost $\sim0.06$ F1) and raised landmark losses.

Untargeted label flipping primarily collapses the attacker’s own model with minimal spillover to other clients; model-level manipulations (e.g., Scaled Boost) inflict stronger local damage and measurable collateral degradation within the attacker’s task group; and attacks that subvert the aggregation stage break task isolation and cause the most severe, system-wide declines, affecting both same-task and cross-task nodes. 

Overall, \textit{ColNet} demonstrates model robustness: it strongly resists isolated label noise and modest manipulations, but remains vulnerable when an attacker either (i) amplifies their aggregate influence or (ii) directly compromises the aggregation routine, suggesting defense priorities of anomaly detection for client updates and hardening aggregation integrity.

\section{Summary and Future Work}
\textit{ColNet} is a decentralized federated multi-task learning framework that groups clients by task, averages backbones within each group, and reconciles group updates via an HCA-style conflict-averse aggregator. An adaptive clustering module discovers and refines group membership over time. Empirical results show consistent gains over alternative aggregators while demonstrating improved robustness. The combination of a shared backbone, task-specific heads, HCA aggregation, and adaptive clustering substantially limits cross-task propagation of client-side attacks.

Future work focuses on fault-tolerant deployment, including leader election and rotation, straggler mitigation, and support for asynchronous rounds, together with stronger aggregation hardening through Byzantine-robust and secure aggregation to balance accuracy, privacy, and robustness. Large-scale distributed evaluations against stronger adaptive adversaries will quantify trade-offs and guide improvements toward comprehensive robustness guarantees.

\bibliographystyle{IEEEtran}
\balance
\bibliography{references}

\end{document}